\documentclass{article}
\usepackage{spconf,amsmath,graphicx}
\usepackage{paralist}
\usepackage{soul}
\usepackage{xcolor}
\usepackage{graphicx}
\usepackage{caption}
\usepackage{subcaption}
\usepackage{flushend}

\usepackage{enumitem}
\setlist{nosep, leftmargin=14pt}

\usepackage{mwe} 

\title{Deep Dirichlet uncertainty for unsupervised out-of-distribution detection of eye fundus photographs in glaucoma screening}

\name{Teresa Ara\'{u}jo$^{1,*}$, Guilherme Aresta$^{1}$\sthanks{T.Araujo and G.Aresta contributed equally to this work. Corresponding author: teresa.safinisterraaraujo@meduniwien.ac.at}, Hrvoje Bogunovi\'c$^{1}$}

\address{Christian Doppler Laboratory for Artificial Intelligence in Retina,\\ Department of Ophthalmology and Optometry, \\Medical University of Vienna, Austria}
\begin{document}
%
\maketitle
%

\begin{abstract}

The development of automatic tools for early glaucoma diagnosis with color fundus photographs can significantly reduce the impact of this disease. However, current state-of-the-art solutions are not robust to real-world scenarios, providing over-confident predictions for out-of-distribution cases. With this in mind, we propose a model based on the Dirichlet distribution that allows to obtain class-wise probabilities together with an uncertainty estimation without exposure to out-of-distribution cases. We demonstrate our approach on the AIROGS challenge, where we achieve a performance similar to other participants without requiring additional annotations or artificially generated out-of-distribution labels.

\end {abstract}


\section{Introduction}


The glaucoma in one of the leading causes of blindness worldwide, with studies suggesting that over 111 million people will suffer from this pathology in 2040~\cite{Tham2014}. The glaucoma's impact can be significantly reduced if diagnosed early~\cite{ViolaStellaMary2016}.
Particularly, eye fundus photography is the preferred approach for glaucoma screening because it is a non-invasive and cost-effective way of assessing the optic disc (OD), where the primary manifestations of the pathology occur. 
However, large scale screening programs are difficult to implement due to the large number of images to be assessed by ophthalmologists.  
Automated systems can improve the success these screening programs by reducing the workload of specialists. Despite the high performance of the current automatic solutions, mostly deep learning (DL) based, these state-of-the-art-systems are usually not robust in real-world scenarios, providing over-confident predictions on out-of-distribution (OOD) cases.

The current DL approaches for glaucoma screening usually follow one of two approaches: 
\begin{inparaenum}[1)]
\item image-wise classification or
\item initial OD detection followed by classification based only on that region of interest~\cite{Orlando2020}.
\end{inparaenum} 
While OD-based approaches tend to have higher performance, they rely on a successful preliminary OD detection, which may fail due to partial motion artifacts or even domain shifts Furthermore they require additional annotations, which makes this type of approaches less transferable to different eye pathologies and consequently resulting in less expandable systems. 

The vast majority of existing classification methods are not robust to OOD cases, providing overconfident predictions and thus limiting the real use of these systems for automated screening. This makes automated OOD estimation an important component of trustworhy AI systems and a trending research topic in the computer vision community. The most generic approaches for OOD detection rely on computing an uncertainty measure \textit{a posteriori} using methods such as Monte Carlo Dropout and models' ensembles, but these approaches are not task-specific, and thus their performance can be subpar. Because of this, other approaches rely on supervised training to perform both classification and OOD detection~\cite{Liang2018,Roy2021}. However, this supervised outlier exposure may lead to an overfit to certain types of OOD cases, which can artificially increase performance in limited test sets without guarantying robustness in real-world applications~\cite{Hsu2020}.
A possible alternative is to use evidential deep learning. For instance, Dirichlet distribution has already been used with success for OOD detection in natural images~\cite{Sensoy2018}. Advantageously, these type of approaches do not need costly ensembles, post-processing or tailored OOD sets to provide a classification uncertainty.  

With this in mind, in this work, we propose an uncertainty-aware deep network that predicts a Dirichlet distribution on the glaucoma class probabilities. During inference, this type of approach allows to obtain class-wise probabilities together with a sample-wise uncertainty $\in[0\,1]$ of that same classification. To the best of our knowledge, this is the first time this approach was applied to medical images. Additionally, to fully automate OOD detection and provide a decision threshold, we rely on the assumption that referable glaucoma detection is only possible if the region of the OD has sufficient image quality for diagnosis. 

\subsection{AIROGS challenge}

We assess our approach on the Artificial Intelligence for RObust Glaucoma Screening Challenge (AIROGS challenge)~\cite{Vente2021}. The main task was to develop an automatic  method for \emph{referable glaucoma} detection in eye fundus image. Additionally, the system should provide a soft and binary decision on whether each image is not diagnosable (\emph{ungradable}), i.e. automatically identify OOD samples and bad quality images. No definition or example of what an ungradable image was provided. 
Furthermore, usage of external fundus image datasets was forbidden. 

\section{Glaucoma classification with uncertainty estimation}

\subsection{Dataset}
The AIROGS development data~\cite{Vente2021data} contains 101\,442 images, from which 3\,270 have referable glaucoma. For our experiments, we randomly split the data into training, validation and test sets with 80\%, 10\% and 10\% of the data, respectively, stratified by class prevalence. Thus, both the validation and the test set contained 10\,145 images, from which 327 were graded as referable glaucoma. All images were resized to the input size of the network.
%
%
The AIROGS test dataset has approximately 11\,000 images. These images and their labels were hidden from the participants, and instead performance evaluation was performed by submitting the algorithm to the AIROGS web platform. However, prior to the final test phase on these images, the challenger organizers allowed to assess the performance of the algorithm on around 10\% of the test data. A total of 3 attempts were possible for this preliminary test phase, and a single attempt for the final test phase.

\subsection{Base architecture} The algorithm was developed using exclusively the AIROGS dataset~\cite{Vente2021data}. 
The classification model is composed of the first two inception blocks from the Inception-V3~\cite{Szegedy2016} network pre-trained on ImageNet~\cite{Russakovsky2015}. Using only these blocks reduces the size of the receptive field which, as it will be addressed later, allows to identify in detail the relevant diagnosis regions and subsequently propose an OOD binary decision.



\subsection{Deep Dirichlet uncertainty estimation}

Our method is based on the direct modeling of the uncertainty following the evidential deep learning approach~\cite{Dempster2008}. In particular, we deal with the $K$ class probabilities as resulting from a Dirichlet distribution, i.e., a belief mass $b_k$ is attributed to each singleton (i.e, class label) $k$, $k\in \{1,...,K\}$, from a set of mutually exclusive singletons, and an overall uncertainty mass $u$ is provided, with $u \geq 0$, $b_k \geq 0$ and $u + \sum_{k=1}^K b_k = 1$. 
Each $b_k$ is computed based on the evidence for that singleton $e_k$ via $b_k = {e_k}/{S}$, where $S$ is the total evidence. 
The prediction uncertainty $u$ is:
\begin{equation}
    u = \frac{K}{S} = \frac{K}{\sum_{k=1}^K(e_k + 1)}.
\label{eq:dir_unc}
\end{equation}
The uncertainty is thus inversely proportional to the total evidence, and in the extreme case of no evidence we have $b_k = 0, \forall k \implies u = 1$. 
This evidence can be modeled by a Dirichlet distribution characterized by $K$ $\alpha_k$ parameters, with $\alpha_k = e_k + 1$.
%
%
%
The probability $\hat{p_k}$ of the class $k$ is given by the mean of the Dirichlet distribution parameters $\hat{p_k} = \frac{\alpha_k}{S}$. 
%

We utilize the uncertainty value $u$ to detect OOD cases.

\subsection{Training loss}

From the network backbone we first obtain $K=2$ logits
, which are clipped to $[-200, 200]$ and then converted to evidences ($e$) using a softplus activation. We train the model with two loss terms based on Kullback-Leibler (KL) divergence. The first term aims at increasing $e$
for the correct class by assessing the divergence between the predicted $\alpha$ and the theoretically maximum $\alpha_\mathrm{max}=201$:

\begin{equation}
L_{\mathrm{KL}_\mathrm{evid}} = \mathrm{KL}\left(D(p_i \rvert {\alpha_i})~||~D(p_i\rvert y_\mathrm{gt}\odot\langle \alpha_\mathrm{max},...,\alpha_\mathrm{max}\rangle)\right)   
\end{equation}
where $y_\mathrm{gt}$ is the reference categorical label. 
A second KL divergence term regularizes the distribution by penalizing the divergence from the uniform distribution in the uncertain cases:
\begin{equation}
L_{\mathrm{KL}_\mathrm{unif}} = \mathrm{KL}\left(D(p_i \rvert \hat{\alpha_i})~||~D(p_i\rvert \langle 1,...,1\rangle)\right)    
\end{equation}
where $D(p_i\rvert \langle 1,...,1\rangle)$ is the uniform Dirichlet distribution and $\hat{\alpha_i}$ is the Dirichlet parameters after removing the non-misleading evidence from the $\alpha_i$ parameters for sample $i$:  $\hat{\alpha_i} = y_i + (1-y_i)\odot \alpha_i$.
The final loss is then defined as: 
 \begin{equation}
     L = L_{\mathrm{KL}_\mathrm{evid}} + a_t L_{\mathrm{KL}_\mathrm{unif}}
 \end{equation}
with $a_t$ being the annealing coefficient that increases as the training progresses. In particular, $a_t = \min(1, t/s)$, where $t$ is the current training epoch and $s$ is the annealing step, gradually increasing the effect of the second term in the loss, avoiding the premature convergence to the uniform distribution for misclassified cases in the beginning of the training~\cite{Sensoy2018}.

\subsection{Training details}

The network receives as input $224 \times 224$ 
pixels RGB images and outputs per sample the glaucoma probability and the confidence of the prediction.
Our model was trained with balanced batches, the data was normalized to the ImageNet standard range and data augmentation included random flips, translations, rotations, scales, Gaussian blur and brightness modifications. The optimizer is Adam with learning rate $10^{-4}$. The method was developed using Python and Tensorflow/Keras 2.5.0 with a NVidia GTX3080 desktop.

\subsection{Out-of-distribution binary decision}

The challenge required participants to indicate, both with a continuous score and a binary label, if an image is ungradable. Since no examples of ungradable images were provided, we made the assumption that diagnosis is only possible if the OD has enough image quality for diagnosis, as glaucoma main structural manifestation occurs in that region.  Thus, we artificially created OOD images by occluding the regions of the images where their Grad-CAM \cite{Selvaraju2017} was greater than 0.5. This allowed us to produce in-distribution (ID) and OOD samples in our validation set, with which we computed the threshold for the binary ungradability decision.  In particular, we contructed a receiver operating characteristic (ROC) curve using $u_{\text{ID}}$ and $u_{\text{OOD}}$. The ROC curve was used for selecting two decision thresholds, one at 0.5 sensitivity ($u=0.35$) and the other at the optimal operating point that minimizes the distance to the (0,1) point ($u=0.13$). We tested both thresholds on the preliminary test phase, and we kept $u=0.35$ as it performed better on that data.

\subsection{Baseline method}

The underlying hypothesis of our method is that the predicting the Dirichlet distribution allows for a better OOD detecion than out-of-the-box classification networks without degrading the classification performance. With this in mind, we compared our solution with a standard Maximum Softmax Probability (MSP), a common baseline for OOD detection~\cite{Roy2021}. In particular, we use the same base architecture with a softmax activation. The remaining design decisions, including training and OOD threshold decision scheme, were kept the same.

\subsection{Evaluation}

The challenge participants were evaluated using: 1) the partial area under the ROC curve (90-100\% specificity) for referable glaucoma (pAUC), 2) sensitivity at 95\% specificity (TPR@95), 3) Cohen's kappa score between the reference and the model's decisions on image ungradability ($\kappa$) and 4) the ungradability AUC (gAUC). 

\begin{figure}[htb]

\begin{minipage}{1.0\linewidth}
  \centering
  \centerline{\includegraphics[width= \linewidth]{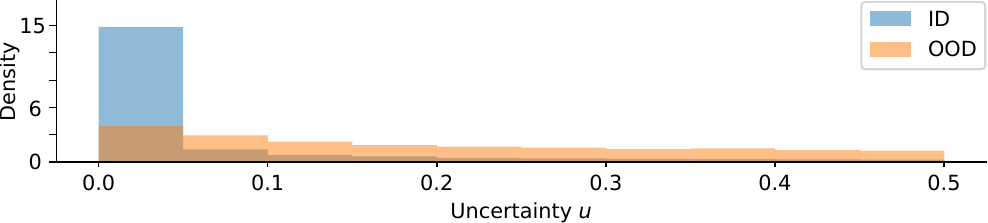}}
  \centerline{(a) Softmax}\medskip
\end{minipage}
\begin{minipage}{1.0\linewidth}
  \centering
  \centerline{\includegraphics[width=\linewidth]{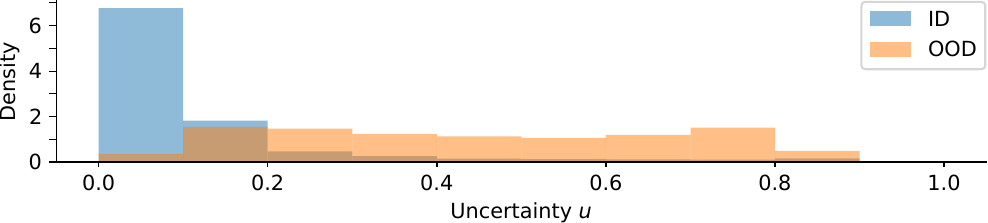}}
  \centerline{(b) Dirichlet}
\end{minipage}
\vspace{-0.5em}
\caption{Uncertainty histogram of the in-distribution (ID) and the artificial out-of-distribution (OOD) cases for the softmax baseline and the proposed approach.}
\label{fig:uncertainty_histogram}
\end{figure}

\section{Results}

%
%
%

\begin{figure}[htb]
    \centering
    
    \begin{minipage}{1.0\linewidth}
  \centering
  \centerline{\includegraphics[width= \linewidth]{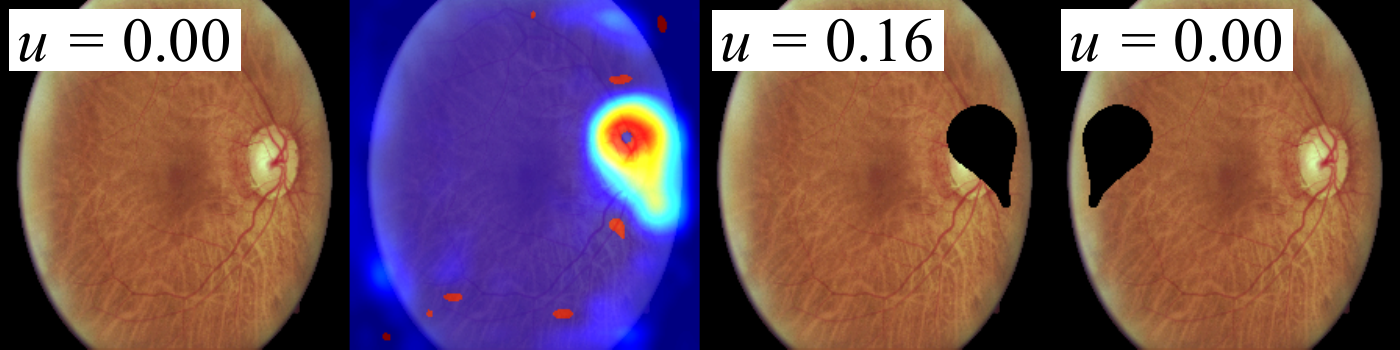}}
  \centerline{(a) Softmax}\medskip
\end{minipage}

\begin{minipage}{1.0\linewidth}
  \centering
  \centerline{\includegraphics[width= \linewidth]{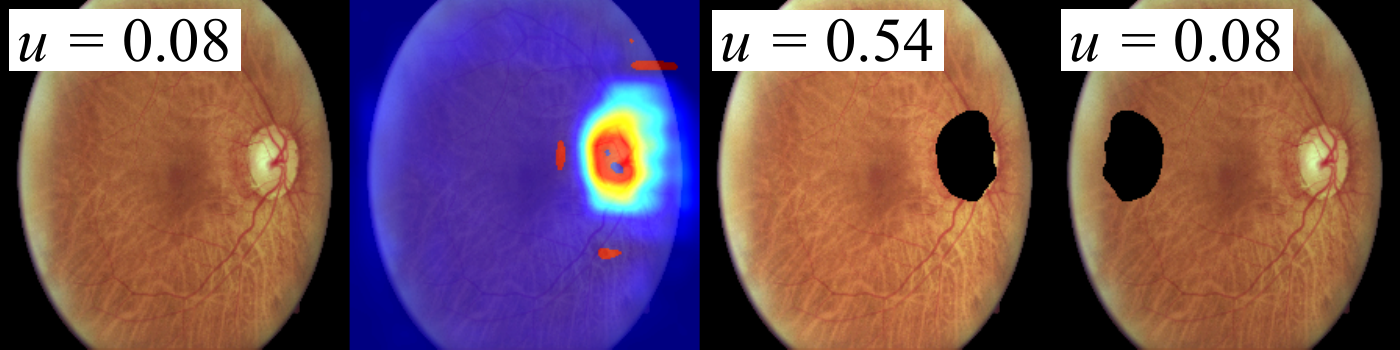}}
  \centerline{(b) Dirichlet}\medskip
\end{minipage}
\vspace{-0.5em}
\caption{Representative example of (left-to-right) original image, Grad-CAM overlay, out-of-distribution by optic disc obscuring, and out-of-distribution by flipping the binarized Grad-CAM, by using the softmax baseline, and the proposed approach, with the corresponding prediction uncertainty $u$.}
    \label{fig:example}
\end{figure}

The uncertainty histogram (Fig.~\ref{fig:uncertainty_histogram}) for ID and OOD shows that the predicted uncertainty $u$ is a viable metric to identify images where the OD is not visible. 
To ensure that this behaviour was due to the OD being obscured, we compared the AUC values for detecting our OOD cases with the values for detecting the cases where the corresponding Grad-CAM mask was flipped vertically, as exemplified in Fig.~\ref{fig:example}. Specifically, for the Dirichlet network, the  achieved AUC values were 0.905 and 0.506, respectively, thus validating our hypothesis that the OD image quality is pivotal for this task. 
Also, directly modeling the Dirichlet distribution instead of using the softmax allows for a more bimodal histogram, and consequently an easier determination of the OOD decision threshold. For instance, in the example on Fig.~\ref{fig:example} the OD occlusion leads to a much higher uncertainty variation for the Dirichlet network in comparison to the standard softmax.  Indeed, as shown in Table~\ref{tab:performance}, in our test dataset, both MSP and our approach have similar performance in all tasks, except for the $\kappa$, which is $18\%$ higher, suggesting that the Dirichlet-based approach allows for a better OOD detection.


\begin{figure*}[htb]
\begin{minipage}{0.3\linewidth}
  \centering
\centerline{\includegraphics[width=\linewidth]{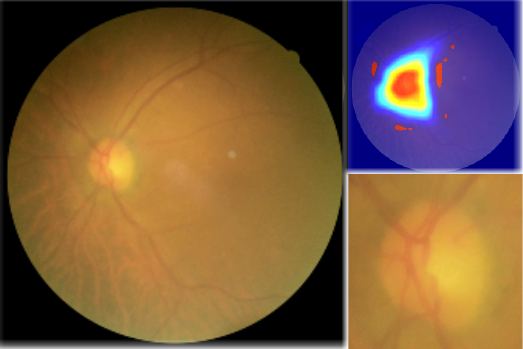}}
  \centerline{(a) RL: no glauc., $p=$0.03, $u=$0.06}\medskip
\end{minipage}
\hfill
\begin{minipage}{0.3\linewidth}
  \centering
  \centerline{\includegraphics[width=\linewidth]{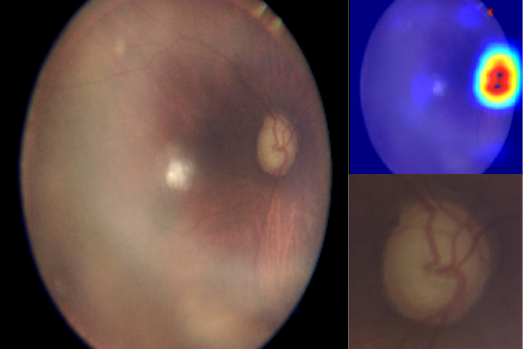}}
  \centerline{(b) RL: glauc., $p=$0.98, $u=$0.04.}\medskip
\end{minipage}
\hfill
\begin{minipage}{0.3\linewidth}
  \centerline{\includegraphics[width=\linewidth]{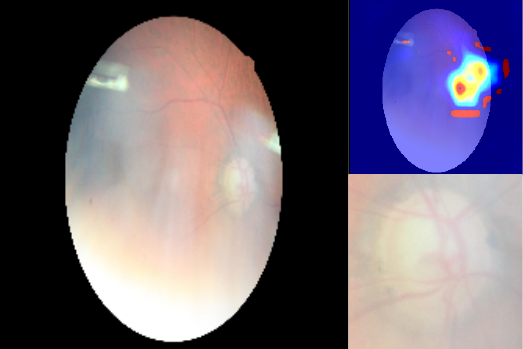}}
  \centerline{(c)  RL: glauc., $p=$0.78, $u=$0.43.}\medskip
\end{minipage}
\begin{minipage}{0.3\linewidth}
  \centering
  \centerline{\includegraphics[width=\linewidth]{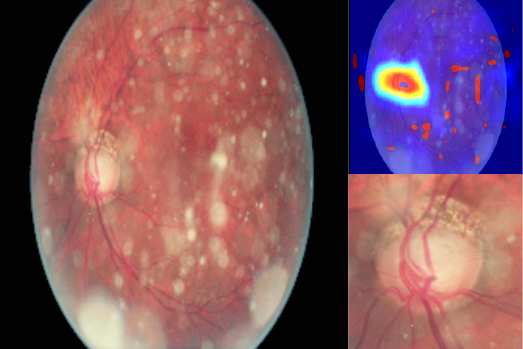}}
  \centerline{(d)  RL: glauc., $p=0.94$, $u=$0.11.}\medskip
\end{minipage}
\hfill
\begin{minipage}{0.3\linewidth}
  \centering
  \centerline{\includegraphics[width=\linewidth]{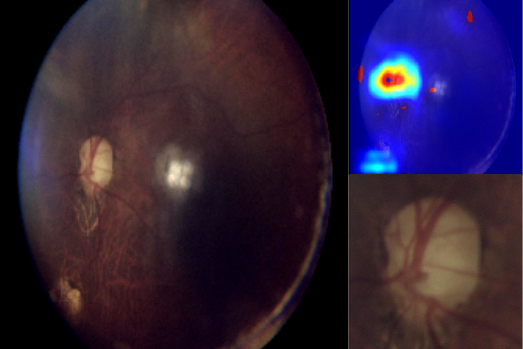}}

  \centerline{(e) RL: glauc., $p=$0.95., $u=$0.09.}\medskip
\end{minipage}
\hfill
\begin{minipage}{0.3\linewidth}
  \centering
  \centerline{\includegraphics[width=\linewidth]{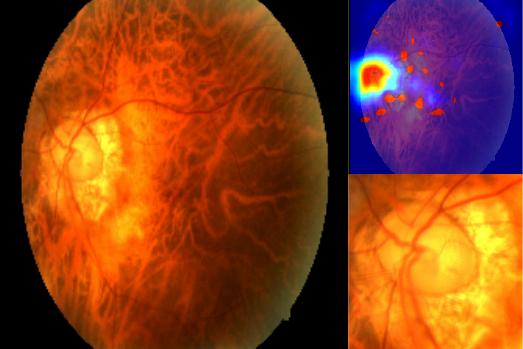}}

  \centerline{(f)  RL: glauc., $p=$0.28, $u=$0.51.}\medskip
\end{minipage}
\caption{Examples of predictions of the proposed model on the test set used in the development phase, together with the Grad-CAM and optic disc detail. RL: reference label; $p$: predicted glaucoma probability; $u$: prediction uncertainty}
\label{fig:images_examples}
\end{figure*}
\begin{table}[htb]
    \centering
    \caption{Performance on our test split, the preliminary and final test phases of the challenge. For comparison, the $\kappa$ on our test set is computed using the optimal operating point of the simulated out-of-distribution detection ROC curves. pAUC: partial AUC (90-100\% specificity) for referable glaucoma; TPR@95: sensitivity at 95\% specificity; $\kappa$: Cohen's kappa score for ungradability; gAUC: ungradability AUC. \label{tab:performance}}
    
    \begin{tabular}{ccccccc} \hline
         \textbf{Method} & \textbf{Test} & \textbf{pAUC} & \textbf{TPR@95} &  $\mathbf{\kappa}$ & \textbf{gAUC}  \\ \hline
         MSP& Ours & 0.91 & 0.88 & 0.58 & 0.88  \\ \hline
         Ours& Ours & 0.92 & 0.90 & 0.69 & 0.90  \\ \hline \hline
         Ours&Prelim. & 0.85 & 0.78 & 0.4452 & 0.87 \\ \hline 
         Ours& Final & 0.84 & 0.75 & 0.51 & 0.90 \\ \hline 
    \end{tabular}
    \label{tab:my_label}
\end{table}

Table~\ref{tab:performance} reports our performance also on the test set of the AIROGS challenge. As shown, besides a 10\% performance drop at TPR@95 and an over-optimistic estimation of $\kappa$,  which were expected given the reduced number of glaucoma cases and complexity of the ungradability task, our approach shows a similar behaviour across the datasets
. Importantly, the model shows high scores across all the metrics without relying on any type of OOD supervision. Indeed, not establishing \textit{a priori} an OOD definition allows the model to leverage both image quality (namely OD visibility) and anatomical variability to estimate the prediction uncertainty. For instance,
%
%
%
Fig.~\ref{fig:images_examples} shows predictions of proposed method on some examples of our test set. From these examples one can see that not only images with good quality are well classified and with low uncertainty (Fig.~\ref{fig:images_examples}-a)), but also overall bad quality images can be correctly predicted with low uncertainty by our method (Fig.~\ref{fig:images_examples}-b), \ref{fig:images_examples}-d) and \ref{fig:images_examples}-e)). Indeed, the image from Fig.\ref{fig:images_examples}-b) shows significant blur, the one from Fig.\ref{fig:images_examples}-d) presents several bright artifacts, and Fig.~\ref{fig:images_examples}-e) shows very low luminosity. However, the OD region, the most relevant for glaucoma diagnosis, is still visible in these examples, and thus the network is able to make predictions with low uncertainty. In contrast, image on Fig.\ref{fig:images_examples}-c), which OD structures are not as visible due to high brightness, was predicted by the network with high uncertainty. 
Fig.~\ref{fig:images_examples}-f) shows an example where an uncommon glaucoma manifestation (and consequently under-represented on the training data) is also identified as an outlier, despite not presenting low quality on the OD region. We believe that this is a desirable behaviour of the system, as it would require a human expert to validate the prediction on a rare pathological case. Overall, these qualitative results suggest that the proposed network is using relevant image information for the diagnosis and that the uncertainty estimation reflects the ability of predicting the correct diagnosis instead of being a simple measure of overall image quality.

\section{Conclusion}
In this paper, we presented our method for the AIROGS challenge which, being based on the Dirichlet distribution, allows to obtain a probability of referable glaucoma and the corresponding prediction uncertainty. Even without explicit supervision, the model is capable of detecting OOD cases while maintain a high performance on the classification task, showing potential for the development of a robust automatic glaucoma screening system.


\section{Acknowledgments}
\label{sec:acknowledgments}

The financial support by the the Christian Doppler Research Association, Austrian Federal Ministry for Digital and Economic Affairs, the National Foundation for Research, Technology and Development, and Heidelberg Engineering is gratefully acknowledged.

\bibliographystyle{IEEEbib}


\begin{thebibliography}{10}

\bibitem{Tham2014}
Yih~Chung Tham, Xiang Li, Tien~Y. Wong, Harry~A. Quigley, Tin Aung, and
  Ching~Yu Cheng,
\newblock ``{Global prevalence of glaucoma and projections of glaucoma burden
  through 2040: A systematic review and meta-analysis},''
\newblock {\em Ophthalmology}, vol. 121, no. 11, pp. 2081--2090, 2014.

\bibitem{ViolaStellaMary2016}
M.C. Viola Stella~Mary, E.B. Rajsingh, and G.R. Naik,
\newblock ``{Retinal Fundus Image Analysis for Diagnosis of Glaucoma: A
  Comprehensive Survey},''
\newblock {\em IEEE Access}, vol. 4, 2016.

\bibitem{Orlando2020}
José~Ignacio Orlando, Huazhu Fu, João Barbossa~Breda, Karel van Keer,
  Deepti~R. Bathula, Andrés Diaz-Pinto, Ruogu Fang, Pheng~Ann Heng, Jeyoung
  Kim, Joon~Ho Lee, Joonseok Lee, Xiaoxiao Li, Peng Liu, Shuai Lu, Balamurali
  Murugesan, Valery Naranjo, Sai Samarth~R. Phaye, Sharath~M. Shankaranarayana,
  Apoorva Sikka, Jaemin Son, Anton van~den Hengel, Shujun Wang, Junyan Wu,
  Zifeng Wu, Guanghui Xu, Yongli Xu, Pengshuai Yin, Fei Li, Xiulan Zhang, Yanwu
  Xu, and Hrvoje Bogunovi{\'{c}},
\newblock ``{REFUGE Challenge: A unified framework for evaluating automated
  methods for glaucoma assessment from fundus photographs},''
\newblock {\em Medical Image Analysis}, vol. 59, 2020.

\bibitem{Liang2018}
Shiyu Liang, Yixuan Li, and R.~Srikant,
\newblock ``{Enhancing the reliability of out-of-distribution image detection
  in neural networks},''
\newblock {\em 6th International Conference on Learning Representations, ICLR
  2018 - Conference Track Proceedings}, pp. 1--15, 2018.

\bibitem{Roy2021}
Abhijit~Guha Roy, Jie Ren, Shekoofeh Azizi, Aaron Loh, Vivek Natarajan, Basil
  Mustafa, Nick Pawlowski, Jan Freyberg, Yuan Liu, Zach Beaver, Nam Vo, Peggy
  Bui, Samantha Winter, Patricia MacWilliams, Greg~S. Corrado, Umesh Telang,
  Yun Liu, Taylan Cemgil, Alan Karthikesalingam, Balaji Lakshminarayanan, and
  Jim Winkens,
\newblock ``{Does Your Dermatology Classifier Know What It Doesn't Know?
  Detecting the Long-Tail of Unseen Conditions},''
\newblock {\em ArXiv}, 2021.

\bibitem{Hsu2020}
Yen~Chang Hsu, Yilin Shen, Hongxia Jin, and Zsolt Kira,
\newblock ``{Generalized ODIN: Detecting Out-of-Distribution Image without
  Learning from Out-of-Distribution Data},''
\newblock {\em Proceedings of the IEEE Computer Society Conference on Computer
  Vision and Pattern Recognition}, pp. 10948--10957, 2020.

\bibitem{Sensoy2018}
Murat Sensoy, Lance Kaplan, and Melih Kandemir,
\newblock ``{Evidential deep learning to quantify classification
  uncertainty},''
\newblock in {\em 32nd Conference on Neural Information Processing Systems
  (NeurIPS 2018)}, Montr{\'{e}}al, 2018.

\bibitem{Vente2021}
Coen~de Vente, Koenraad~A. Vermeer, Nicolas Jaccard, Bram~van Ginneken, Hans~G. Lemij and Clara~I. S{\'{a}}nchez,
\newblock ``{AIROGS: Artificial Intelligence for RObust Glaucoma Screening
  Challenge},'' 2021.

\bibitem{Vente2021data}
Coen~de Vente, Koenraad~A. Vermeer, Nicolas Jaccard, Bram~van Ginneken, Hans~G. Lemij and Clara~I. S{\'{a}}nchez,
\newblock ``{Rotterdam EyePACS AIROGS train set},'' .

\bibitem{Szegedy2016}
Christian Szegedy, Vincent Vanhoucke, Sergey Ioffe, Jonathon Shlens, and
  Zbigniew Wojna,
\newblock ``{Rethinking the Inception Architecture for Computer Vision},''
\newblock {\em Proceedings of the IEEE Computer Society Conference on Computer
  Vision and Pattern Recognition (CVPR)}, pp. 2818--2826, 2016.

\bibitem{Russakovsky2015}
Olga Russakovsky, Jia Deng, Hao Su, Jonathan Krause, Sanjeev Satheesh, Sean Ma,
  Zhiheng Huang, Andrej Karpathy, Aditya Khosla, Michael Bernstein,
  Alexander~C. Berg, and Li~Fei-Fei,
\newblock ``{ImageNet Large Scale Visual Recognition Challenge},''
\newblock {\em International Journal of Computer Vision}, vol. 115, no. 3, pp.
  211--252, 2015.

\bibitem{Dempster2008}
Arthur~P. Dempster,
\newblock ``{A Generalization of Bayesian Inference},''
\newblock in {\em Classic Works of the Dempster-Shafer Theory of Belief
  Functions}, pp. 73--104. Springer Berlin Heidelberg, Berlin, Heidelberg,
  2008.

\bibitem{Selvaraju2017}
Ramprasaath~R. Selvaraju, Michael Cogswell, Abhishek Das, Ramakrishna Vedantam,
  Devi Parikh, and Dhruv Batra,
\newblock ``{Grad-CAM: Visual Explanations from Deep Networks via
  Gradient-Based Localization},''
\newblock in {\em 2017 IEEE International Conference on Computer Vision
  (ICCV)}, Venice, 10 2017, pp. 618--626, IEEE.

\end{thebibliography}

\end{document}